\documentclass[10pt,twocolumn,letterpaper]{article}

\usepackage[pagebackref=true,breaklinks=true,letterpaper=true,colorlinks,bookmarks=false]{hyperref}

\usepackage{cvpr}

\usepackage{times}
\usepackage{epsfig}
\usepackage{graphicx}
\usepackage{amsmath}
\usepackage{amssymb}
\usepackage{makecell}
\usepackage{bm}
\usepackage{color}
\usepackage[scientific-notation=true]{siunitx}

 \cvprfinalcopy 

\ifcvprfinal\fi
\begin{document}

\title{Bidirectional Multirate Reconstruction for Temporal Modeling in Videos}

\author{Linchao Zhu\hspace{2em}Zhongwen Xu\hspace{2em}Yi Yang\\
	University of Technology Sydney\\
	{\tt\small \{zhulinchao7, zhongwen.s.xu, yee.i.yang\}@gmail.com } 
}
\maketitle
\thispagestyle{empty}

\begin{abstract}
Despite the recent success of neural networks in image feature learning,
a major problem in the video domain is the lack of sufficient labeled data for learning to model temporal information.
In this paper, we propose an unsupervised temporal modeling method that learns from untrimmed videos.
The speed of motion varies constantly, \eg, a man may run quickly or slowly.
We therefore train a Multirate Visual Recurrent Model (MVRM) by encoding frames of a clip with different intervals.
This learning process makes the learned model more capable of dealing with motion speed variance.
Given a clip sampled from a video, we use its past and future neighboring clips as the temporal context, and reconstruct the two temporal transitions,
\ie, present$\rightarrow$past transition and present$\rightarrow$future transition, reflecting the temporal information in different views.
The proposed method exploits the two transitions simultaneously by incorporating a bidirectional reconstruction which consists of a backward reconstruction and a forward reconstruction.
We apply the proposed method to two challenging video tasks, \ie, complex event detection and video captioning,
in which it achieves state-of-the-art performance.
Notably, our method generates the best single feature for event detection with a relative improvement of 10.4\% on the MEDTest-13 dataset and
achieves the best performance in video captioning across all evaluation metrics on the YouTube2Text dataset.
\end{abstract}


\section{Introduction}
Temporal information plays a key role in video representation modeling. In earlier years,
hand-crafted features, \eg, Dense Trajectories (DT) and improved Dense Trajectories (iDT)~\cite{wang2011action,wang2013action},
use local descriptors along trajectories to model video motion structures.
Despite achieving promising performance, DT and iDT are very expensive to extract, due to the heavy computational cost of optical flows and
it takes about a week to extract iDT features for 8,000 hours of web videos using 1,000 CPU cores~\cite{xu2015discriminative}.
Deep visual features have recently achieved significantly better performance in image classification and detection tasks than hand-crafted features
at an efficient processing speed~\cite{krizhevsky2012imagenet,he2015deep,girshickICCV15fastrcnn}.
However, learning a video representation on top of deep Convolutional Neural Networks (ConveNets) remains a challenging problem.
Two-stream ConvNet~\cite{simonyan2014two} is groundbreaking in learning video motion structures over short video clips.
Although it achieves comparable performance to iDT for temporally trimmed videos,
two-stream ConvNet still needs to extract optical flows.
The heavy cost severely limits the utility of methods based on optical flows, especially in the case of large scale video data.

\begin{figure}[t]
	\centering
	\includegraphics[width=0.98\linewidth]{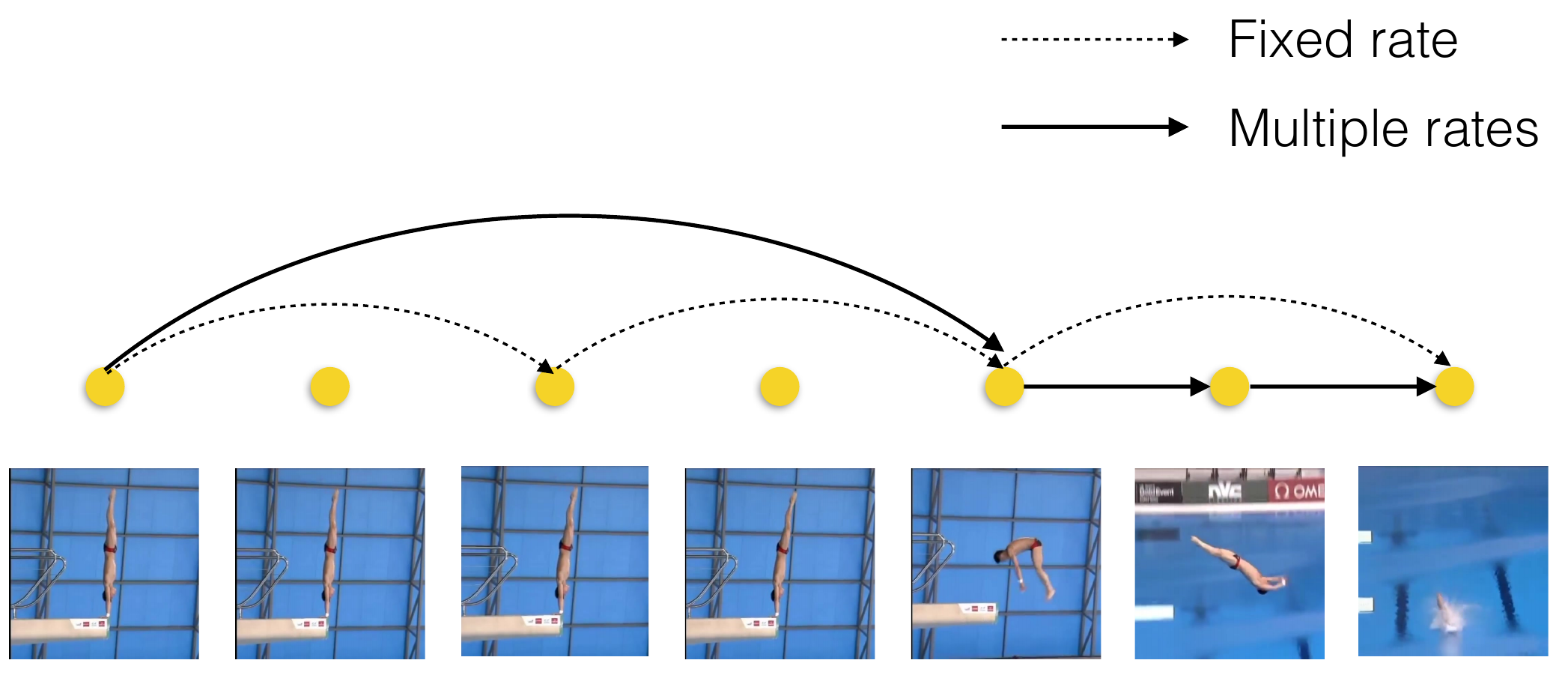}
	\caption{Frame sampling rate should vary in accordance with different motion speed. In this example, only the last three frames have fast motion.
             The dashed arrow corresponds to a fixed sampling rate, while the solid arrow corresponds to multiple rates.
	}
	\label{fig:model_tasks}
\end{figure}

Extending 2D ConvNet to 3D, C3D ConvNet has been demonstrated to be effective for spatio-temporal modeling and it avoids extracting optical flows.
However, it can only model temporal information in short videos, usually of 16 frames~\cite{c3d}.
Recurrent Neural Networks (RNNs), particularly Long Short-Term Memory (LSTM)~\cite{lstm,ng2015beyond} and a modified Hierarchical Recurrent Neural Encoder (HRNE)~\cite{pan2015hierarchical}, have been used
to model temporal information in videos. A major limitation of~\cite{ng2015beyond} and~\cite{pan2015hierarchical} is that the input frames are encoded with a fixed sampling rate when training the RNNs.
On the other hand, the motion speed of videos varies even in the same video.
As shown in the Figure~\ref{fig:model_tasks}, there is almost no apparent motion in the first four frames, but fast motion is observed in the last three frames.
The encoding rate should be correspondingly low for the first four frames, but high for the last three, as indicated by the solid arrow.
The fixed rate strategy, however, is redundant for the first four frames, while important information for the last three frames is lost.
The gap between the fixed encoding rate and motion speed variance in real world videos may degrade performance, especially when the variance is extensive.

Notwithstanding the appealing ability of end-to-end approaches for learning a discriminative feature,
such approaches require a large amount of labeled data to achieve good performance with plausible generalization capabilities.
Compared to images, a large number of videos are very expensive to label by humans.
For example, the largest public human-labeled video dataset (ActivityNet)~\cite{caba2015activitynet} only has 20,000 labeled videos while the ImageNet dataset has over one million labeled instances~\cite{ILSVRC15}.
Temporal ConvNet trained on the UCF-101 dataset~\cite{soomro2012ucf101} with about 10,000 temporally trimmed videos did not generalize well on temporally a untrimmed dataset~\cite{xu2015uts}.
Targeting short video clips, Srivastava~\etal~\cite{icml2015_srivastava} proposed training a composite autoencoder in an unsupervised manner to learn video temporal structures,
essentially by predicting future frames and reconstructing present frames.
Inspired by a recent study on neuroscience which shows that a common brain network underlies the capacity both to remember the past and imagine the future~\cite{schacter2012future},
we consider reconstructing two temporal transitions,
\ie, present$\rightarrow$past transition and present$\rightarrow$future transition.
Importantly, video motion speed changes constantly in untrimmed videos and Srivastava~\etal directly used an LSTM with a single fixed sampling rate, making it vulnerable to
motion speed variance.

In this paper, we propose an unsupervised method to learn from untrimmed videos for temporal information modeling without the heavy cost of computing optical flows. It makes the following two major contributions.
First, our Multirate Visual Recurrent Model adopts multiple encoding rates,
and together with the reading gate and the updating gate in the Gated Recurrent Unit,
it enables communication between different encoding rates and collaboratively learns a multirate representation which is robust to motion speed variance in videos.
Second, we leverage the mutual benefit of two learning processes by reconstructing the temporal context in two directions.
The two learning directions regularize each other, thereby reducing the overfitting problem.
The two contributions yield a new video representation, which achieves the best performance in two different tasks.
Note that the method proposed in~\cite{xu2015discriminative} has been demonstrated to be the best single feature for event detection,
and our method outperforms this method with a relative improvement of 10.4\% and 4.5\% on two challenging datasets, \ie, MEDTest-13 and MEDTest-14 respectively.
In the video captioning task, our single feature outperforms other state-of-the-art methods across all evaluation metrics, most of which use multiple features.
It is worthwhile mentioning that in very rare cases, one method can outperform all others for video captioning over all evaluation metrics.
These results demonstrate the effectiveness of the proposed method.


\section{Related Work}
Research efforts to improve visual representations for videos have been ongoing.
Local features such as HOF~\cite{laptev2008learning} and MBH~\cite{dalal2006human} extracted along spatio-temporal tracklets have been used as motion descriptors in the Dense Trajectories feature~\cite{wang2011action} and its variants~\cite{wang2013action}.
However, it is notoriously inefficient to extract hand-crafted features like improved Dense Trajectories (iDT)~\cite{wang2013action,xu2015discriminative},
mostly due to the dense sampling nature of local descriptors and the time-consuming extraction of optical flows.
On the other hand, the classification performance of state-of-the-art hand-crafted features has been surpassed by many methods based on neural networks in web video classification and action recognition tasks~\cite{xu2015discriminative,WangQT15a}.

\noindent \textbf{Convolutional Networks for video classification}.
One way to use ConvNets for video classification is to perform temporal pooling over convolutional activations.
Ng~\etal~\cite{ng2015beyond} proposed learning a global video representation by using max pooling over the last convolutional layer across video frames.
Wang~\etal~\cite{WangQT15a} aggregated ConvNet features along the tracklets obtained from iDT.
Xu~\etal~\cite{xu2015discriminative} applied VLAD encoding~\cite{JegouDSP10} over ConvNet activations and
found that the encoding methods are superior to mean pooling. The other common solution is to feed multiple frames as input to ConvNets.
Karpathy~\etal~\cite{karpathy2014large} proposed a convolutional temporal fusion network, but it is only marginally better than the single frame baseline.
Tran~\etal~\cite{c3d} avoided the extraction of optical flows by utilizing 3D ConvNets to model motion information. 
Simonyan and Zisserman~\cite{simonyan2014two} took optical flows as the flow image input to a ConvNet, and this two-stream
network has much better performance than the previous networks on action recognition.

\noindent\textbf{Recurrent Networks for video classification}.
Ng~\etal~\cite{ng2015beyond} and Donahue~\etal~\cite{donahue2015long} investigated the modeling of temporal structures in videos with Long Short-Term Memory (LSTM)~\cite{lstm}.
However, even with five-layer LSTMs, trained on millions of videos, they do not show promising performance compared to ConvNets~\cite{ng2015beyond}.
Patraucean~\etal~\cite{PatrauceanHC16} used a spatio-temporal autoencoder to
model video sequences through optical flow prediction and reconstruction of the next frame.
Ballas~\etal~\cite{ballas2015delving} used a Convolutional Gated Recurrent Unit (ConvGRU) which leverage information from different spatial levels of the activations.
Srivastava~\etal~\cite{icml2015_srivastava} used LSTM to model video sequences in an unsupervised way.
In this work, we utilize the RNNs on video representation learning, improving the representation by being aware of the multirate nature of video content.
Moreover, the temporal consistency between frames in the neighborhood is incorporated into the networks in an unsupervised way,
providing richer training information and creating opportunities to learn from abundant untrimmed videos.

\noindent\textbf{Video captioning}. Video captioning has emerged as a popular task
in recent years, since it bridges visual understanding and natural language description. Conditioned on the visual context, RNNs produce one word per step to generate captions for videos.
Venugopalan~\etal~\cite{s2vt} used a stacked sequence to sequence (seq2seq)~\cite{sutskever2014sequence} model, in which an LSTM is used as a video sequence encoder and the other LSTM serves as a caption decoder.
Yao~\etal~\cite{yao2015capgenvid} incorporated the temporal attention mechanism in the description decoding stage.
Pan~\etal~\cite{pan2015hierarchical} proposed using a hierarchical LSTM to model videos sequences, while Yu~\etal~\cite{yu2015video}
used a hierarchical GRU network to model the structure of captions.
In this work, we demonstrate that the strong video representation learned in our model improves the video captioning task, confirming the generalization ability of our features.


\section{Multirate Visual Recurrent Models}
In this section, we introduce our approach for video sequence modeling.
We first review the structure of Gated Recurrent Unit (GRU) and extend the GRU to a multirate version.
The model architecture for unsupervised representation learning is then introduced, which is followed
by task specific models for event detection and video captioning. In the model description, we omit all bias terms in order to increase readability.

\subsection{Multirate Gated Recurrent Unit}
\label{sec:gru}
\noindent\textbf{Gated Recurrent Unit.} At each step $t$, a GRU cell takes a frame representation $\mathbf{x}_t$
and previous state $\mathbf{h}_{t-1}$ as inputs and generates
a hidden state $\mathbf{h}_t$ and an output $\mathbf{o}_t$ which are calculated by,
\begin{equation}\begin{aligned}
    \mathbf{r}_t &= \sigma(\mathbf{U}_{r} \mathbf{x}_t + \mathbf{V}_{r} \mathbf{h}_{t-1}), \\
     \mathbf{z}_t &= \sigma(\mathbf{U}_{z} \mathbf{x}_t + \mathbf{V}_{z} \mathbf{h}_{t-1}), \\
     \bar{\mathbf{h}_t} &= \tanh(\mathbf{U}_{\bar{h}} \mathbf{x}_t + \mathbf{V}_{\bar{h}}(\mathbf{r}_t \odot \mathbf{h}_{t-1})), \\
     \mathbf{h}_t &= (1 - \mathbf{z}_t) \odot \mathbf{h}_{t-1} + \mathbf{z}_t \odot \bar{\mathbf{h}_{t}}, \\
     \mathbf{o}_t  &= \mathbf{W}_o\mathbf{h}_t,
\label{eq:model_gru}
\end{aligned}\end{equation}
where $\mathbf{x}_t$ is the input, $\mathbf{r}_t$ is the reset gate, $\mathbf{z}_t$ is the update gate,
$\mathbf{h}_t$ is the proposed state, $\bar{\mathbf{h}_t}$ is the internal state,
$\sigma$ is the sigmoid activation function, $\mathbf{U_{\ast}}$ and $\mathbf{V}_{\ast}$ are weight matrices,
and $\odot$ is element-wise multiplication.
The output $\mathbf{o}_t$ is calculated by a linear transformation from the state $\mathbf{h}_t$.
We denote the whole process as:

\begin{equation}
\mathbf{h}_t, \mathbf{o}_t = \text{GRU}(\mathbf{x}_t, \mathbf{h}_{t-1}),
\end{equation}
and when it has iterated $S$ steps, we can obtain the state of the last step $\mathbf{h}_S$.

\noindent\textbf{Multirate Gated Recurrent Unit (mGRU)}.
Inspired by clockwork RNN~\cite{koutnik2014clockwork}, we extend the GRU cell to a multirate version.
The clockwork RNN uses delayed connections for inputs
and inter-connections between steps to capture longer dependencies.
Unlike traditional RNNs where all units in the states follow the protocol in Eq.~\ref{eq:model_gru},
states and weights in the clockwork RNN are divided into groups to model information at different rates.
We divide state $\mathbf{h}_t$ into $k$ groups, and each group $g_i$
has a clock period $T_i$, where $i \in \{1, \ldots, k\}$.
$T_i$ can be arbitrary numbers, and we empirically use $k=3$ and set $T_1, T_2, T_3 = 1, 3, 6$.
Faster groups (with smaller $T_i$) take inputs more frequently than slower groups, and the slower module skips more inputs.
Formally, at each step $t$, matrices of the group satisfying $(t~\verb|MOD|~T_i) = 0$ are activated and are used to calculate
the next state, which is
\begin{equation}\begin{aligned}
    \mathbf{r}_t^i &= \textstyle{\sigma(\mathbf{U}_{r}^i \mathbf{x}_t + \sum_{j=1}^{k}{\mathbf{V}_{r}^{i,j} \mathbf{h}_{t-1}^j})}, \\
    \mathbf{z}_t^i &= \textstyle{\sigma(\mathbf{U}_{z}^i \mathbf{x}_t + \sum_{j=1}^{k}{\mathbf{V}_{z}^{i,j} \mathbf{h}_{t-1}^j})}, \\
    \bar{\mathbf{h}_t^i} &= \textstyle{\tanh(\mathbf{U}_{\bar{h}}^i \mathbf{x}_t + \sum_{j=1}^{k}{\mathbf{V}_{\bar{h}}^{i,j}(\mathbf{r}_t^i \odot \mathbf{h}_{t-1}^j)})}, \\
    \mathbf{h}_t^i &= \textstyle{(1 - \mathbf{z}_t^i) \odot \mathbf{h}_{t-1}^i + \mathbf{z}_t^i \odot \bar{\mathbf{h}_{t}^i}}, 
\end{aligned}\end{equation}
where the state weight matrices $\mathbf{V_{\ast}}$ are divided into $k$ block-rows and each block-row is
partitioned into $k$ block-columns. $\mathbf{V}_{\ast}^{i,j}$ denotes the sub-matrix in block-row $i$ and block-column $j$.
The input weight matrices $\mathbf{U}_{\ast}$ are divided $k$ block-rows and $\mathbf{U}_{\ast}^{i}$ denotes the weights
in block-row $i$ and
\begin{gather}
    \label{eq:fast_slow_mode}
    \textstyle{\sum_{j=1}^{k}{\mathbf{V}_{\ast}^{i,j}\mathbf{h}_{t-1}^{j}}} = \begin{cases}
        \sum_{j=1}^{i}{\mathbf{V}_{\ast}^{i,j}\mathbf{h}_{t-1}^j}, \ \ \ \ \ \text{Fast $\rightarrow$ slow mode} \\
        \sum_{j=i}^{k}{\mathbf{V}_{\ast}^{i,j}\mathbf{h}_{t-1}^j}, \ \ \ \ \ \text{Slow $\rightarrow$ fast mode}
    \end{cases}
\end{gather}

\begin{figure}[t]
\centering
\includegraphics[width=0.87\linewidth]{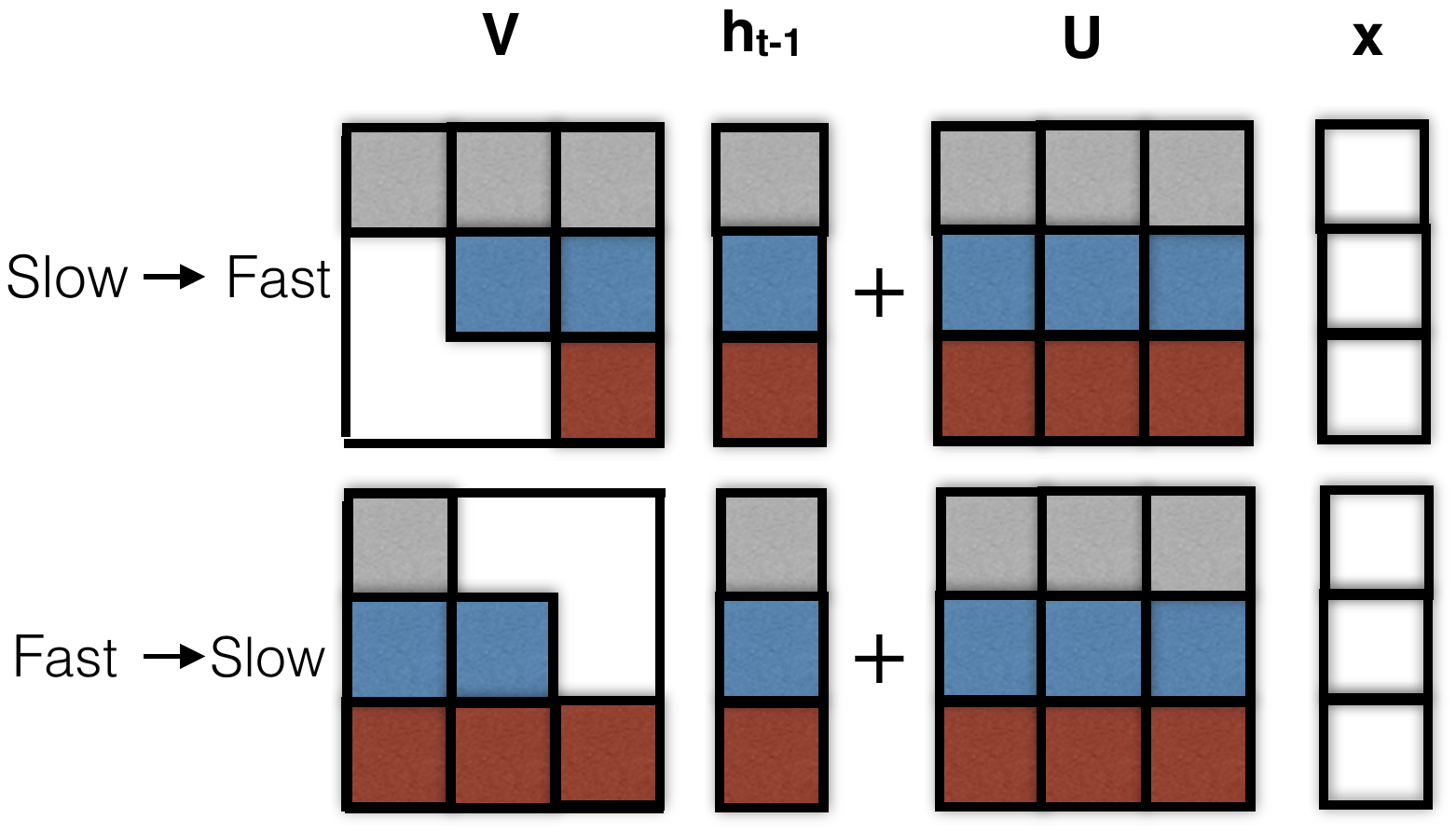}
    \caption{We illustrate the two modes in the mGRU. In the slow to fast mode, the state matrices $\mathbf{V}_{\ast}$
    are block upper-triangular matrices and in the fast to slow mode, they are block lower-triangular matrices.}
\label{fig:cwrnn_matrix}
\end{figure}

Two modes can be used for state transition.
In the slow to fast mode, states of faster groups consider previous slower states,
thus the faster states incorporate information not only at the current speed but
also information that is slower and more coarse.
The intuition for the fast to slow mode is that when the slow mode is activated,
it can take advantage of the information already encoded in the faster states.
The two modes are illustrated in Figure~\ref{fig:cwrnn_matrix}.
Empirically, we use the fast to slow mode in our model as it performed better in the preliminary experiments.

If $(t\ \verb|MOD|~T_i) \neq 0$, the previous state is directly passed over to the next state,
\begin{gather}
    \mathbf{h}_t^i = \mathbf{h}_{t-1}^i.
\end{gather}

\begin{figure}[t]
\centering
\includegraphics[width=0.87\linewidth]{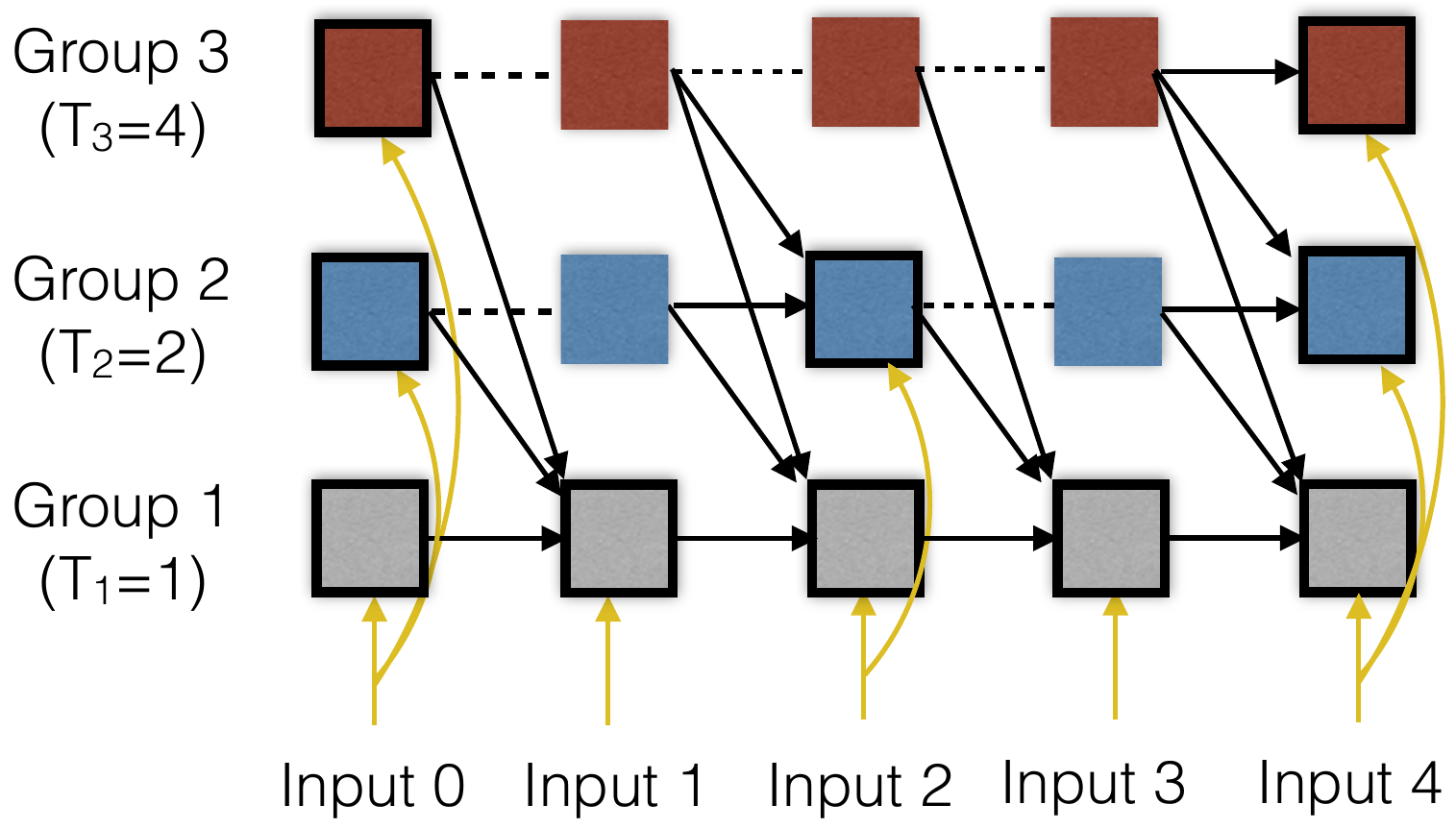}
\caption{Unrolled mGRU.
    In the example, the state is divided into three groups and the slow to fast mode is shown.
    At each step $t$,
         groups satisfying $(t~\texttt{MOD}~T_i) = 0$ are activated (cells with black border). For example,
         at step 2, group 1 and group 2 are activated. The activated groups take the frame input and previous
         states to calculate the next states. For those that are inactivated, we simply pass the previous states
         to the next step. Group 1 is the fastest and group 3 is the slowest with larger $T_i$. The slow to fast mode is the mode by which
         the slower groups pass the states to the faster groups.
    }
\label{fig:cwrnn}
\end{figure}

Figure~\ref{fig:cwrnn} illustrates the state iteration process.
Note that not all previous modules are considered to calculate the next state at each step, thus fewer parameters will be used
and the training will be more efficient.

\subsection{Unsupervised Video Sequence Reconstruction}
\label{sec:model_unsup}
Video sequences are highly correlated to their neighboring context clips.
We use the idea of context reconstruction for video sequence modeling.
The similar methods have been successfully applied for language modeling and other language tasks~\cite{mikolov2013efficient,kiros2015skip}.
In the unsupervised training process, we follow the classic sequence-to-sequence (seq2seq) model~\cite{sutskever2014sequence} where
an encoder encodes a sequence of inputs and passes the last state to the decoder for target sequence generation.
In our scenario, the mGRU encoder takes frame-level features extracted from the pre-trained convolutional models as inputs
and generates the state at each step which will be attended by the decoders.
The state of the last step of the encoder is passed to the decoder, \ie, $\mathbf{h}_{0}^{\text{dec}} = \mathbf{h}_S^{\text{enc}}$.
Two decoders are used to predict the context sequences of the inputs, \ie, reconstructing the frame-level representations of the previous sequence and next sequence.

\begin{figure}[t]
\centering
\includegraphics[width=0.98\linewidth]{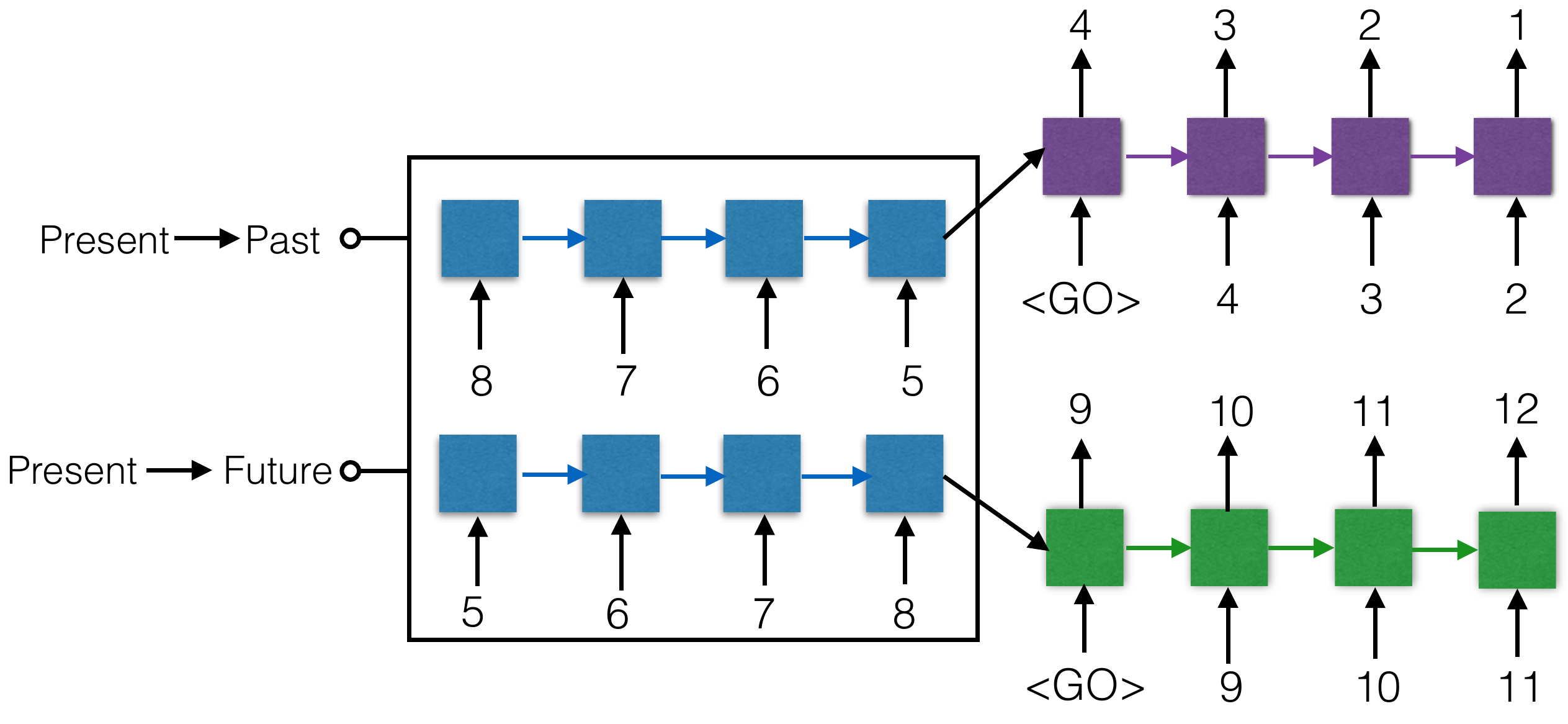}
\caption{The model architecture of unsupervised video representation learning.
         In this model, two decoders are used to predict surrounding contexts by reconstructing
         previous frames and next frame sequences.
         The ``\texttt{<GO>}'' input, which is a zero vector, is used at step 0 in the decoder.
         During training, one of the two decoders is used with a probability of 0.5 for reconstruction.}
\label{fig:unsup_model_arch}
\end{figure}

\noindent\textbf{Decoder}. We use the seq2seq model with attention mechanism to model video temporal structures via context reconstruction.
We denote that $\mathbf{Y}=(\mathbf{y}_1, \mathbf{y}_2, \ldots, \mathbf{y}_n)$ is the previous sequence of input sequence $\mathbf{X}$,
and $\mathbf{Z}=(\mathbf{z}_1, \ldots, \mathbf{z}_n)$ is the next sequence.
The decoder is a GRU conditioned on the encoder outputs $\mathbf{o}_{1,\ldots,S}^{\text{enc}}$
and the last step state $\mathbf{h}_S^{\text{enc}}$ of the encoder.
We use the attention mechanism at each step to help the decoder to decide which frames in the input sequence might be related to the next frame reconstruction.
At step $t$, the decoder $\phi$ generates the prediction $\mathbf{o}_t^{\text{dec}}$ by calculating,
\begin{equation}\begin{aligned}
    \mathbf{y}_t^{\text{attn}} &= \text{Linear}(\mathbf{y}_t, \mathbf{a}_{t-1}), \\
    \mathbf{h}_{t}^{\text{dec}}, \mathbf{o}_{t}^{\text{attn}} &= \text{GRU}(\mathbf{y}_t^{\text{attn}}, \mathbf{h}_{t-1}^{\text{dec}}), \\
    e_t^{i} &= \mathbf{v}^{\text{T}}\text{tanh}(\mathbf{W}_{he}\mathbf{h}_t^{\text{dec}} + \mathbf{W}_{oe}\mathbf{o}_i^{\text{enc}}), \\
    a_{t}^{i} &= \textstyle{{\exp({e}_t^{i})}\mathbin{/}{\sum_{{j=1}}^{S}{\exp({e}_t^{j})}}}, \\
    \mathbf{a}_t &= \textstyle{\sum_{{i=1}}^{S}{a_t^{i}\mathbf{o}_i^{\text{enc}}}}, \\
    \mathbf{o}_t^{\text{dec}} &= \text{Linear}(\mathbf{o}_t^{\text{attn}}, \mathbf{a}_t)
    \label{eq:model_decoder},
\end{aligned}\end{equation}
where
$\text{Linear}(\mathbf{a}, \mathbf{b}) = \mathbf{W}_a\mathbf{a} + \mathbf{W}_b\mathbf{b}$,
$a_t^i$ is the normalized attention weight for encoder output $\mathbf{o}_i^{\text{enc}}$ and $\mathbf{a}_t$ is the weighted average of the encoder outputs.
We use two decoders that do not share parameters: one for the past sequence reconstruction and the other for the future sequence reconstruction~(Figure~\ref{fig:unsup_model_arch}).
The decoders are trained to minimize the reconstruction loss of two sequences, which is
\begin{gather}
    \begin{aligned}
        &\sum_t{\ell(\phi({\mathbf{y}_{<t}, \mathbf{o}_{1,\ldots,S}^{\text{enc}}, \mathbf{h}_S^{\text{enc}}}; \mathbf{\theta}), \mathbf{y}_t)}+ \\
        &\sum_{t'}{\ell(\phi(\mathbf{z}_{<t'}, \mathbf{o}_{1,\ldots,S}^{\text{enc}}, \mathbf{h}_{S}^{\text{enc}}; \mathbf{\theta'}), \mathbf{z}_{t'})}.
    \end{aligned}
\end{gather}

We choose the Huber loss for regression due to its robustness following Girshick~\cite{girshickICCV15fastrcnn},
\begin{gather}
    \ell(y, \bar{y}) = \begin{cases}
        \dfrac{1}{2}(y - \bar{y})^2  \ \ \ \ \ \ \ \ \ \ \ \ \ \ \text{for} \left|y-\bar{y}\right| \le \delta, \\
        \delta\left|y-\bar{y}\right| - \dfrac{1}{2}\delta^2  \ \ \ \ \ \text{otherwise}.
    \end{cases}
\end{gather}
We set $\delta=0.5$ in all experiments.

For the past reconstruction, we reverse the input order as well as the target order
to minimize information lag~\cite{sutskever2014sequence}.
The two decoders are trained with the encoder via backpropagation,
and we regularize the network by randomly dropping one decoder for each batch.
As we have two decoders in our model, each decoder will
have the probability of being chosen for training of 0.5 (Figure.~\ref{fig:unsup_model_arch}).

During unsupervised training, we uniformly sample video frames and extract frame-level features from convolutional models.
We set the sequence length to $K$, \ie, the encoder takes $K$ frames as inputs,
while the decoders reconstruct previous $K$ frames and next $K$ frames.
We randomly sample a temporal window of consecutive $3K$ frames (3 segments) during training. If the video length is less
than $3K$, we pad zeros for each segment.

\subsection{Complex Event Detection}
\label{sec:model_med}
We validate the unsupervised learned features on the task of complex event detection.
We choose the TRECVID Multimedia Event Detection (MED) task as it is more dynamic and complex compared to
the action recognition task, in which the target action duration is short and usually lasts only seconds.
As the features from the unsupervised training are not discriminative, \ie, label information has not been applied during training, we further train the encoder for video classification.
We use the mGRU encoder to encode the video frames and take the last hidden state in the encoder for classification.
We do not apply losses at each step, \eg, the LSTM model in~\cite{ng2015beyond}, as the video data in our task is untrimmed, which is more
noisy and redundant.
We use the network structure of FC(1024)-ReLU-Dropout(0.5)-FC(1024)-ReLU-Dropout(0.5)-FC(\textit{class\_num}+1)-Softmax.
Since there are background
videos which do not belong to any target events, we add another class for these videos.

During supervised training, we first initialize the weights of the encoder with the weights pre-trained via unsupervised context reconstruction.
For each batch, instead of uniformly sampling videos within the training set,
we keep the ratio of the number of positive and background videos to $1: 2$.
We bias the mini-batch sampling because of the imbalance between the positive and negative examples.

During inference, the encoder generates multirate states at each step,
and there are several ways to pool the states to obtain a global video representation.
One simple approach is to average the outputs, and the obtained global video representation is then classified with a Linear SVM.
The other way is to encode the outputs with an encoding method.
Xu~\etal~\cite{xu2015discriminative} found that Vector of Locally Aggregated Descriptors (VLAD)~\cite{JegouDSP10}
encoding outperforms average pooling and Fisher Vectors~\cite{perronnin2010improving} over ConvNets activations by a large margin on the MED task.
We thus apply the VLAD encoding method to encode the RNN representations.

Given inputs $\mathbf{X} = \{\mathbf{x}_1, \mathbf{x}_2, \ldots, \mathbf{x}_N\}$
and centers $\mathbf{C}=\{\mathbf{c}_1, \ldots, \mathbf{c}_K\}$ which are calculated by the k-means algorithm on sampled inputs, for each $k \in \{1, \ldots, K\}$, we have,
\begin{equation}
\label{eq:vlad_classic}
    \mathbf{u}_k=\sum_{i:\text{Nearest}(\mathbf{x}_i)=\mathbf{c}_k}\mathbf{x}_i-\mathbf{c}_k,
\end{equation}
where $\mathbf{x}_i$ is assigned to the center $\mathbf{c}_k$ if it is the nearest center.
Concatenating $\mathbf{u}_k$ over all $K$ centers, we obtain the feature vector of size $DK$ where $D$ is the dimension of $\mathbf{x}_i$.
Normalization methods are used to improve the encoding performance.
Power normalization, often signed square rooting (SSR), is usually used to convert each element $x_i$ into
$\text{sign}(x_i)\sqrt{\left|x_i\right|}$.
The intra-normalization method normalizes representations for each center,
followed by the $\ell_2$ normalization for the whole feature vector~\cite{perronnin2010improving}.
The final normalized representation is classified with a Linear SVM.

Note that the states in mGRU are divided into groups, we thus encode the state of the three different scales independently.
We combine the three scores by average fusion.

\subsection{Video Captioning}
\label{sec:model_captioning}
We also demonstrate the generalization ability of our proposed video representation on the video captioning task.
In video captioning, an encoder is used to encode video representations and a decoder is used
to generate video descriptions. We follow the basic captioning decoding process.
Given a video sequence $\mathbf{X}$ and a description sequence
$Y=\{y_1, \ldots, y_N\}$, where each word is represented by a one-hot vector and
a one-of-$K$ ($K$ is the vocabulary size) embedding is used in the decoder input to represent a discrete word with a continuous vector,
the overall objective is to maximize the log-likelihood of the generated sequence,
\begin{equation}
    \max_{\bm{\theta}}{\sum_{t=1}^{N}{\log\Pr(y_t|y_{<t}, \mathbf{X}; \bm{\theta})}}.
\end{equation}
Softmax activation is used on the decoder output to obtain the probability of word $y_t$.
The attention mechanism (Eq.~\ref{eq:model_decoder}) is used in both the input and output of the decoder.


\section{Experiments}
We show the results of our experiments on complex event detection and video captioning tasks.
We implement our model using the TensorFlow framework~\cite{TensorFlow}.
Source code and trained models will be released upon acceptance.

\subsection{Complex Event Detection}
\subsubsection{Dataset}
We collect approximately 220,000 videos without label information from TRECVID MED data, which excludes
videos in MEDTest-13 and MEDTest-14, for unsupervised training.
The average length of the collected videos is 130 seconds with a total duration of more than 8,000 hours.

We use the challenging MED datasets with labels, namely, TRECVID MEDTest-13 100Ex~\cite{med13} and
TRECVID MEDTest-14 100Ex~\cite{med14} for video classification\footnote{Development data is not updated for TRECVID MED 15 and TRECVID MED 16 competition.}.
There are 20 events in each dataset, 10 of which overlap.
It consists of approximately 100 positive exemplars for each event in the training set,
and 5,000 negative exemplars. In the testing set, there are about 23,000 videos and
the total duration in each collection is approximately 1,240 hours.
The average video length is 120 seconds.
These videos are temporally untrimmed YouTube videos of various resolutions and quality.
We use the mean Average Precision (mAP) as the performance metric
following the NIST standard~\cite{med13,med14}.

\subsubsection{Model Specification}
For both unsupervised training and classification, we uniformly sample video frames at the rate of 1 FPS
and extract features for each frame from GoogLeNet with the Batch Normalization~\cite{ioffe2015batch} pre-trained on ImageNet.
Following standard image preprocessing procedures, the shorter edges of frames are rescaled to 256 and we crop the image to $224\times224$.
We use activations after the last pooling layer and obtain representations with length 1,024.
There are 20 classes in the MEDTest-13 and MEDTest-14 datasets, thus with the background class, we have 21 classes in total.
In the training stage, we set sequence length $K$ to 30 and pad zeros if the video has fewer than 30 frames.
During inference, we take the whole video as input and use 150 steps.

\noindent\textbf{Training details}. We use the following settings in all experiments unless otherwise stated.
The model is optimized with ADAM~\cite{kingma2014adam},
and we fix the learning rate at \num{1e-4} and clip the global gradients at norm 10.
We use a single RNN layer for both the encoder and decoder, and the cell size is set to 1,024.
We set the attention size to 50 and regularize the network by using Dropout~\cite{srivastava2014dropout}
in the input and output layer~\cite{pham2014dropout}.
We also add Dropout when the decoder copy state from the encoder and all dropout probability is set to 0.5.
Weights are initialized with Glorot uniform initialization~\cite{glorot2010understanding} and weight decay of \num{0.0001} is applied for regularization.

In the supervised training, we initialize the weights of the encoder using the learned weights during unsupervised learning,
and the same sequence length is used as in the unsupervised training stage.

\subsubsection{Results}
\noindent\textbf{Average pooling}. For the GoogLeNet baseline, we average frame-level features and use a Linear SVM for classification.
For our model, we first train an unsupervised encoder-decoder model with mGRU and fine-tune the encoder
with label information.
To make a fair comparison with the GoogLeNet baseline, we extract outputs of the mGRU encoder at each step and average them to obtain a global representation for classification.
Note that both feature representations have same dimensions and we empirically set $C=1$ for both of the linear classifiers.
The result is shown in Table~\ref{expr_cw_vs_gru} and shows that our model with temporal structure learning
is able to encode valuable temporal information for classification.
\begin{table}
\makegapedcells
\setcellgapes{1pt}
\begin{center}\begin{tabular}{|l|c|c|c|}
\hline
    Methods                   &   MEDTest-13     &     MEDTest-14 \\
\hline\hline
    GoogLeNet                 &   32.0           &     25.1       \\
\hline
    mGRU                    &   \textbf{39.6}           &     \textbf{32.2}       \\
\hline
\end{tabular}\end{center}
\caption{Comparison between GoogLeNet features and our mGRU model.
         Average pooling is used for both models.
         The result shows our feature representation significantly outperforms the GoogLeNet feature.
         }
\label{expr_cw_vs_gru}
\end{table}

\noindent\textbf{VLAD Encoding}. We now show that VLAD encoding is useful for aggregating RNN representations.
We compare our method with GoogLeNet features using VLAD encoding.
Following~\cite{xu2015discriminative}, we set the number of k-means centers to 256 and the dimension of PCA is 256.
Three scales are learned at each step for our mGRU model.
We divide the state into three segments and each sub-state is individually aggregated by VLAD.
Note that each encoded representation has the same feature vector length as the GoogLeNet model, and we use late fusion to combine the scores of the three scales.
The results in Table~\ref{expr_VLAD} show that our mGRU model outperforms GoogLeNet features when encoded by VLAD.
It also shows that VLAD encoding outperforms average pooling for RNN representations.
Our model also achieves state-of-the-art performance on the MEDTest-13 and MEDTest-14 100Ex datasets.
\begin{table}
\makegapedcells
\setcellgapes{1pt}
\begin{center}\begin{tabular}{|l|c|c|}
\hline
    Methods                                           &   MEDTest-13 &   MEDTest-14 \\
\hline\hline
    \makecell{GoogLeNet}                              &   42.0           &    33.6         \\
\hline
    \makecell{mGRU}                                 &   \textbf{44.5}           &    \textbf{37.3}         \\
\hline
\end{tabular}\end{center}
\caption{Comparison between GoogLeNet and mGRU models when
         VLAD encoding is used to aggregate frame-level representations.}
\label{expr_VLAD}
\end{table}

\subsubsection{Ablation Study}

We compare several variants in the unsupervised training, and show the performance of different components.
The results are shown in Table~\ref{expr_our_own_cmp}.
We obtain features from the unsupervised model by extracting states
from the encoder at each step, which are then averaged to obtain a global video representation.
The results show that the representation learning from unsupervised training without discriminative information also achieves good results.

\noindent \textbf{Attention}. We compare our model with a model without the attention mechanism, where temporal attention is not used
and the decoder is forced to perform reconstruction based only on the last encoder state, \ie, ``mGRU w/o attention'' in Table~\ref{expr_our_own_cmp}.
The results show that the attention mechanism
is important for learning good video representations and also helps the learning process of the encoder.

\noindent \textbf{Context}. In a model without context reconstruction, \ie, only one decoder is used (autoencoder),
neither past nor future context information is considered, \ie, ``mGRU w/o context'' in Table~\ref{expr_our_own_cmp}.
The results show that with context prediction, the encoder has to consider temporal information around the video clip,
which models the temporal structures in a better way.

\noindent \textbf{Multirate}. We also show the benefit of using mGRU by comparing it with the basic GRU, \ie, ``mGRU w/o multirate'' in Table~\ref{expr_our_own_cmp}. Note that the mGRU model has fewer parameters but
better performance. It shows that an mGRU that encodes multirate video information is capable of learning better representations
from long, noisy sequences.

\noindent \textbf{Pre-training}. We now show the advantages of the unsupervised pre-training process by comparing an encoder with random initialization
with the same encoder whose weights are initialized by the unsupervised model.
The result is shown in Table~\ref{expr_pretrain} and demonstrates that the unsupervised training process
is beneficial to video classification. It incorporates context information in the encoder, which
is an important cue for the video classification task.

\begin{table}
\makegapedcells
\setcellgapes{1pt}
\begin{center}\begin{tabular}{|c|c|c|c|}
\hline
    Methods                             &   MEDTest-13 &   MEDTest-14 \\
\hline\hline
    mGRU w/o attention                  &   32.7           &     27.5       \\
\hline
    mGRU w/o context                    &   37.1           &     30.1       \\
\hline
    mGRU w/o multirate                  &   36.5           &     29.3       \\
\hline
    mGRU (full)                         &   37.4           &     30.6       \\
\hline
\end{tabular}\end{center}
\caption{Comparison between mGRU and other variants in the unsupervised training stage. Detailed discussion can be found in text.}
\label{expr_our_own_cmp}
\end{table}

\begin{table}
\makegapedcells
\setcellgapes{1pt}
\begin{center}\begin{tabular}{|l|c|c|}
\hline
    Methods                                           &   MEDTest-13 &   MEDTest-14 \\
\hline\hline
    \makecell{mGRU \\ (random)}                     &    38.3          &   29.5          \\
\hline
    \makecell{mGRU \\ (pre-trained)}                 &    39.6          &   32.2          \\
\hline
\end{tabular}\end{center}
\caption{Comparison between models which have the same structure but different initialization.
    This shows that good initialization enables better features to be learned.}
\label{expr_pretrain}
\end{table}


\subsubsection{Comparison with the State-of-the-art}
We compare our model with other models and the results are shown in Table~\ref{expr_med_comp}.
Our single model achieves the state-of-the-art performance on both the MEDTest-13 and MEDTest-14 100Ex settings
compared with the performances of other single models.
We report the C3D result by using the pre-trained model~\cite{c3d} and we set the length of the input short clip to 16.
Features are averaged across clips which are classified with a Linear SVM.
Our model with VLAD encoding outperforms previous state-of-the-art results with 4.2\% on MEDTest-13 100Ex and
1.6\% on MEDTest-14 100Ex.

\begin{table}[t]
\makegapedcells
\setcellgapes{1pt}
\begin{center}\begin{tabular}{|c|c|c|c|c|}
\hline
    Models                                                           &  MEDTest-13 & MEDTest-14  \\
\hline\hline
    IDT + FV~\cite{xu2015discriminative}                               &   34.0      &     27.6    \\
\hline
    IDT + skip + FV~\cite{lan2015beyond}                                 &   36.3      &     29.0    \\
\hline
    VGG + RBF~\cite{zha2015exploiting}                           &    -        &     35.0    \\
\hline
    \makecell{C3D~\cite{c3d} \\ \footnotesize{(Our implementation)}}                                                  &   36.9  &   31.4 \\
\hline
    VGG16 + VLAD~\cite{xu2015discriminative}                            &    -        &     33.2    \\
\hline
    \makecell{VGG16 + LCD + \\VLAD~\cite{xu2015discriminative}}                      &   40.3      &     35.7    \\
\hline
    \makecell{LSTM autoencoder~\cite{icml2015_srivastava}\\ \footnotesize{(Our implementation)}} &  38.2 &   31.0 \\
\hline
    \makecell{GoogLeNet + VLAD \\ \footnotesize{(Our implementation)}} &   42.0      &     33.6    \\
\hline\hline
    Our method                                                              &   \textbf{44.5}      &     \textbf{37.3}    \\
\hline
\end{tabular}\end{center}
\caption{Comparison with other methods. We achieve state-of-the-art performance on both MEDTest-13 and MEDTest-14 100Ex datasets.}
\label{expr_med_comp}
\end{table}


\subsection{Video Captioning}
We now validate our model on the video captioning task. Our single model outperforms
previous state-of-the-art single models across all metrics.

\subsubsection{Dataset}
We use the YouTube2Text video corpus~\cite{chen2011collecting} to evaluate our model on the video
captioning task. The dataset has 1,970 video clips with an average duration of 9 seconds.
The original dataset contains multi-lingual descriptions covering various domains,
\eg, sports, music, animals.
Following~\cite{venugopalan:naacl15}, we use English descriptions only and
split the dataset into training, validation and testing sets containing 1,200, 100, 670
video clips respectively.
In this setting, there are 80,839 descriptions in total with about 41 sentences per video clip.
The vocabulary size we use is 12,596 including \verb|<GO>|, \verb|<PAD>|, \verb|<EOS>|, \verb|<UNK>|.

\subsubsection{Evaluation}
We evaluate the performance of our method on the test set using the evaluation script provided by~\cite{chen2015microsoft}
and the results are returned by the evaluation server.
We report BLEU~\cite{papineni2002bleu}, METEOR~\cite{denkowski2014meteor} and CIDEr~\cite{vedantam2015cider} scores
for comparison with other models.
We stick with a single rule during model selection, namely we choose the model
with the highest METEOR score on the validation set.

\subsubsection{Model Specification}
The video length in the YouTube2Text dataset is short, thus we uniformly sample
frames at a higher frame rate of 15 FPS.
The sequence length is set to 50 and we use the default hyper-parameters in the last experiment.
We use two different convolutional features for the video captioning task, \ie, GoogLeNet features and
ResNet-200 features~\cite{he2016identity}.
We use beam search during decoding by default and set the beam size to 5 following~\cite{yu2015video} in all experiments.
Attention size is set to 100 empirically.

\subsubsection{Results}
We first use GoogLeNet features and the result is shown in Table~\ref{expr_msvd_inception_v2}.
We compare our mGRU with GRU which shows that mGRU outperforms GRU on all metrics except BLEU@1.
However, the difference is only 0.04\%.
We initialize the mGRU encoder via unsupervised context learning and the result shows
that with good initialization, performance is improved by more than 1.0\% on the BLEU and CIDEr scores and 0.6\% on the METEOR score
compared with random initialization.
We also utilize the recent ResNet-200 network as a convolutional model. We use the pre-trained model
and follow the same image preprocessing method. The result of using ResNet-200 is shown in Table~\ref{expr_msvd_resnet}
and demonstrates that our MVRM method not only works better than GRU on different tasks, but also works better on different convolutional models.
Additionally, we can improve all the metrics with ResNet-200 network.

\begin{table}
\small
\begin{center}\begin{tabular}{|l|c|c|c|c|c|c|}
\hline
    Methods                                                  &  B@1     & B@2     &  B@3    &  B@4      &  M    &  C      \\
\hline\hline
    \footnotesize{GRU}                                       &  79.46   & 67.52   &  57.98  &    47.14  & 32.31 &  72.46   \\
\hline
    \footnotesize{mGRU}                                    &  79.42   & 67.79   &  58.32  &    48.12  & 32.79 &  73.21  \\
\hline
    \footnotesize{\makecell[l]{mGRU+\\pre-train}}           &  \textbf{80.76}   & \textbf{69.49}   &  \textbf{60.03}  &    \textbf{49.45}  & \textbf{33.39} &  \textbf{75.45}       \\
\hline
\end{tabular}\end{center}
    \caption{Comparison between different models on YouTube2Text dataset. GoogLeNet features are used as frame-level representations.
            \textbf{B}, \textbf{M}, \textbf{C} are short for BLEU, METEOR, CIDEr.}
\label{expr_msvd_inception_v2}
\end{table}

\begin{table}
\small
\begin{center}\begin{tabular}{|l|c|c|c|c|c|c|}
\hline
    Methods                                                  &  B@1      & B@2       &  B@3    &  B@4      &  M       &  C      \\
\hline\hline
    \footnotesize{GRU}                                       & 80.88     & 70.15     &  61.08  &  51.06    & 33.48   & 79.16   \\
\hline
    \footnotesize{mGRU}                                    & 82.03     & 71.41     &  62.38  &  52.49    & 33.91    & 78.41   \\
\hline
    \footnotesize{\makecell[l]{mGRU+\\pre-train}}          &  \textbf{82.49}     &  \textbf{72.16}     &  \textbf{63.30}   &  \textbf{53.82}     &  \textbf{34.45}    &  \textbf{81.20} \\
\hline
\end{tabular}\end{center}
\caption{Comparison between different models on YouTube2Text dataset. ResNet-200 features are used as frame-level representations.
        \textbf{B}, \textbf{M}, \textbf{C} are short for BLEU, METEOR, CIDEr.}
\label{expr_msvd_resnet}
\end{table}

\subsubsection{Comparison with the State-of-the-art}
We compare our methods with other models on the YouTube2Text dataset. Results are shown in Table~\ref{expr_msvd_camp}.
``S2VT''~\cite{s2vt} is the first model to use a general encoder-decoder model for video captioning.
``Temporal Attention''~\cite{yao2015capgenvid} uses the temporal attention mechanism on the video frames to obtain better
results. ``Bi-GRU-RCN''~\cite{yao2015capgenvid} uses a ConvGRU to encode activations from different convolutional layers.
``LSTM-E''~\cite{pan2015jointly} uses embedding layers to jointly project visual and text features.
Our MVRM method has similar performance to~\cite{pan2015hierarchical}, but with the pre-training stage, we outperform~\cite{pan2015hierarchical} in all metrics.
Some methods fuse additional motion features like C3D~\cite{c3d} features, \eg, Pan~\etal~\cite{pan2015hierarchical} obtained 33.9\% on METEOR after combing multiple features.
With ResNet-200, we can obtain \textbf{34.45\%} on METEOR.

\begin{table}
\begin{center}\begin{tabular}{|l|c|c|c|c|c|c|}
\hline
    Methods                                                                                 &  BLEU@4 &  METEOR &  CIDEr \\
\hline\hline
    \footnotesize{S2VT~\cite{s2vt}}                                                         &  -      &  29.20   &   -    \\
\hline
    \footnotesize{Temporal attention~\cite{yao2015capgenvid}}                               &  41.92  &  29.60  &  51.67 \\
\hline
    \footnotesize{\makecell[l]{GoogLeNet+\\Bi-GRU-RCN$_1$~\cite{ballas2015delving}}}           &  48.42  &  31.70  &  65.38 \\
\hline
    \footnotesize{\makecell[l]{GoogLeNet+\\Bi-GRU-RCN$_2$~\cite{ballas2015delving}}}           &  43.26  &  31.60  &  68.01 \\
\hline
    \footnotesize{VGG+LSTM-E}~\cite{pan2015jointly}                                         &  40.20   &  29.50   &   -    \\
\hline
    \footnotesize{C3D+LSTM-E}~\cite{pan2015jointly}                                         &  41.70   &  29.90   &   -    \\
\hline
    \footnotesize{\makecell[l]{GoogLeNet+HRNE+\\Attention~\cite{pan2015hierarchical}}}      &  43.80   &  33.10   &   -    \\
\hline
    \footnotesize{VGG+p-RNN~\cite{yu2015video}$^{\ast}$}                                         &  44.30   &  31.10   &   62.10   \\
\hline
    \footnotesize{C3D+p-RNN~\cite{yu2015video}$^{\ast}$}                                         &  47.40   &  30.30   &  53.60   \\
\hline\hline
    \footnotesize{\makecell[l]{GoogLeNet+MVRM}}                               &  \textbf{49.45}  & \textbf{33.39}   & \textbf{75.45}  \\
\hline
\end{tabular}\end{center}
\caption{Comparison with other models without fusion.
         $^\ast$ denotes that the model is trained with different settings (\cite{yu2015video} used the train+val data for training).
         }
\label{expr_msvd_camp}
\end{table}


\section{Conclusion}
In this paper, we propose a Multirate Visual Recurrent Model to learn multirate representations for videos.
We model the video temporal structure via context reconstruction, and show that unsupervised
training is important for learning good representations for both video classification and video captioning.
The proposed method achieves state-of-the-art performance on two tasks.
In the future, we will investigate the generality of the video representation in other challenging tasks,
\eg, video temporal localization~\cite{caba2015activitynet} and video question answering~\cite{zhu2015uncovering,tapaswi2015movieqa}

{\small
\bibliographystyle{ieee}
\bibliography{skip}
}

\end{document}